\documentclass[conference]{IEEEtran}
\IEEEoverridecommandlockouts

\usepackage{cite}
\usepackage{amsmath,amssymb,amsfonts}
\usepackage{algorithmic}
\usepackage{graphicx}
\usepackage{textcomp}
\usepackage{xcolor}
\usepackage{tabularx}
\usepackage{booktabs}
\usepackage{enumitem}
\usepackage{xcolor}
\usepackage{subcaption}

\def\BibTeX{{\rm B\kern-.05em{\sc i\kern-.025em b}\kern-.08em
    T\kern-.1667em\lower.7ex\hbox{E}\kern-.125emX}}
\begin{document}

\title{Accuracy is Not Agreement: Expert-Aligned Evaluation of Crash Narrative Classification Models\\
}
\author{
Sudesh Bhagat\textsuperscript{1,*}, Ibne Farabi Shihab\textsuperscript{2,*}, Anuj Sharma\textsuperscript{1} \\
\textsuperscript{1}Department of Civil, Construction and Environmental Engineering, Iowa State University, Ames, IA, USA \\
\textsuperscript{2}Department of Computer Science, Iowa State University, Ames, IA, USA \\
\textsuperscript{*}Equal contribution
}

\maketitle

\begin{abstract}
This study investigates the relationship between deep learning (DL) model accuracy and expert agreement in classifying crash narratives. We evaluate five DL models—including BERT variants, USE, and a zero-shot classifier—against expert labels and narratives, and extend the analysis to four large language models (LLMs): GPT-4, LLaMA 3, Qwen, and Claude. Our findings reveal an inverse relationship: models with higher technical accuracy often show lower agreement with human experts, while LLMs demonstrate stronger expert alignment despite lower accuracy. We use Cohen’s Kappa and Principal Component Analysis (PCA) to quantify and visualize model-expert agreement, and employ SHAP analysis to explain misclassifications. Results show that expert-aligned models rely more on contextual and temporal cues than location-specific keywords. These findings suggest that accuracy alone is insufficient for safety-critical NLP tasks. We argue for incorporating expert agreement into model evaluation frameworks and highlight the potential of LLMs as interpretable tools in crash analysis pipelines.
\end{abstract}


\section{Introduction}

Road traffic crashes remain a major global public health issue, with U.S. fatal crashes rising 9.9\% in 2021 despite vehicle safety improvements \cite{Chowdhury2023}. To better understand crash dynamics, researchers increasingly rely on crash databases that combine structured data with unstructured crash narratives—free-text descriptions from responding officers that offer contextual insight often missing from coded fields. These narratives have been effectively used to study various crash types, including work zones \cite{Swansen2013}, military collisions \cite{Pollack2013}, and distraction-related incidents \cite{Dube2016}. Prior work also shows that narrative-derived features can uncover hidden risk factors \cite{zhang2013risk} and support scenario reconstruction \cite{najm2004development}. Manual review of narratives, however, is labor-intensive and subjective, limiting scalability. To address this, researchers have increasingly adopted natural language processing (NLP) and deep learning (DL) for classification tasks \cite{Sayed2021}. While video-based multimodal approaches show promise \cite{shihaba2024leveraging,shihab2025crash}, text-based models remain more feasible due to lower resource demands and broad availability of narrative data \cite{zou2018visualization}. Despite their technical promise, DL models often lack interpretability and can produce outputs that diverge from expert judgment, which is especially problematic in safety-critical contexts. Studies in work zones and bicycle-related crashes \cite{Swansen2013, McKnight2003, Lopez2022} show that high accuracy alone does not ensure practical utility. Moreover, narrative accuracy is limited by reporting inconsistencies from officers or data entry errors \cite{Sayed2021, Elvik2015}. This gap between model performance and expert interpretation poses a risk to real-world deployment. In this work, we propose an evaluation framework that complements accuracy with expert agreement metrics such as Cohen's Kappa and applies Principal Component Analysis (PCA) to visualize model-expert divergence, offering a more robust validation pathway for automated crash classification systems.

Our key contributions are as follows:

\begin{itemize}
    \item We systematically evaluate agreement between five DL models, four large language models (LLMs), and human expert judgments using Kappa statistics with bootstrap confidence intervals.
    
    \item We employ PCA to visually interpret classification differences and clustering behaviors among models and experts.
    
    \item We highlight the trade-off between high model accuracy and low expert alignment, especially in the case of specialized DL models, and demonstrate that general-purpose LLMs and models like USE, despite lower accuracy, better capture expert reasoning.
    
    \item We argue for integrating expert-informed evaluation pipelines in future DL-based safety applications, potentially using LLMs as cost-effective intermediaries for model validation.
\end{itemize}


\section{Methods}\subsection{Data Collection and Preprocessing}

We extracted key attributes from two primary data sources:
\begin{itemize}
    \item \textbf{Vehicle data}: CRASH\_KEY, DRIVERAGE, SPEEDLIMIT, VCONFIG, DRIVERDIST
    \item \textbf{Crash data}: CRASH\_KEY, COUNTY, CASENUMBER, FIRSTHARM, CRCOMANNER, MAJORCAUSE, DRUGALCREL, WEATHER1, LIGHT, ROADTYPE, CSEVERITY, CRASH\_YEAR, CRASH\_DATE, DRIVERDIST, WZ\_RELATED
\end{itemize}

This study utilizes the Iowa Department of Transportation crash report narratives for 2019-2020. The narratives describe crash events in natural language format, written by law enforcement officials. The average narrative contains 38 words, with a high standard deviation (25.7) indicating significant length variation. The ground-truth labels (intersection-related vs non-intersection-related) were derived from the crash report's coded fields, particularly the "location" attribute. The crash narrative dataset contained a 'crash key' column and five narrative fields (Narrative 1 through Narrative 5), which we combined to generate comprehensive narratives for each crash key. The complete dataset included 100,812 crash data records for 2019 and 72,700 for 2020, 57,479 crash narrative records for 2019 and 36,888 for 2020. The dataset comprised 173,512 crash data records and 94,367 crash narrative records across both years (2019-2020).

To standardize the text, the narrative text fields underwent preprocessing, including tokenization, stop word removal, and lemmatization. The ground truth labels for the intersection/non-intersection classification were derived from the structured data fields in the crash reports, specifically the "Intersection Type" field, which provided a reliable indicator for training our models, as detailed in Table \ref{table:roadtype}.

\begin{table}[!t]
\caption{Variable Levels (Road Type)}
\label{table:roadtype}
\begin{tabularx}{\linewidth}{|c|X|}
\hline
\textbf{Code} & \textbf{Road Type Description} \\
\hline
1 & Non-intersection: Non-junction/no special feature \\
\hline
2 & Non-intersection: Bike lanes \\
\hline
3 & Non-intersection: Railroad grade crossing \\
\hline
4-7 & Other Non-intersection types (Crossover-related) \\
\hline
10 & Intersection: Roundabout \\
\hline
11 & Intersection: Traffic circle \\
\hline
12 & Intersection: Four-way intersection \\
\hline
13-18, 97 & Other Intersection types \\
\hline
20-24, 98 & Interchange-related types \\
\hline
99, 96 & Unknown or Not Reported \\
\hline
\end{tabularx}
\end{table}

\begin{figure}[!t]
    \centering
    \includegraphics[width=\linewidth]{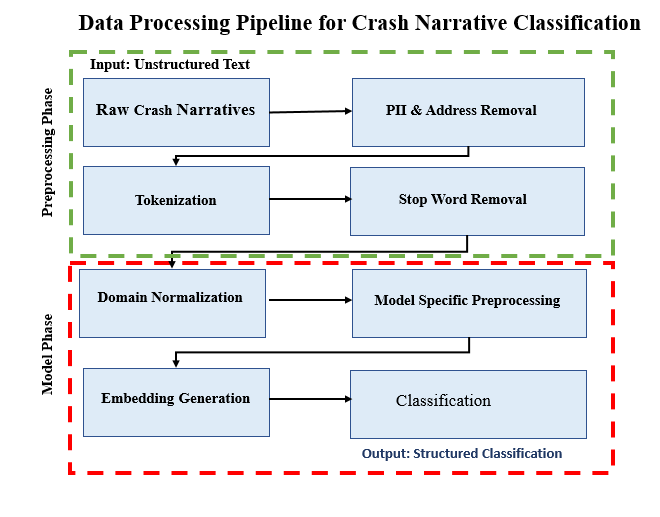}
    \caption{Data Processing Pipeline: Our multi-stage approach for transforming raw crash narratives into structured representations suitable for deep learning classification. Each box represents a processing step that transforms the data sequentially from raw text to model-ready formats.}
    \label{fig:processing}
\end{figure}

Figure \ref{fig:processing} outlines the data processing pipeline used to convert raw crash narratives into structured inputs for model training. The narratives, which contained PII, addresses, stop words, and alphanumeric data (e.g., license plates), underwent thorough cleaning to preserve privacy and enhance model accuracy. Preprocessing included PII removal, customized tokenization and stop word filtering, and normalization of domain-specific terms and abbreviations.

\subsection{Expert Annotations}
We assembled a panel of four domain experts with extensive transportation engineering and traffic operations backgrounds to evaluate the alignment between model predictions and human judgment systematically. Each expert had a minimum of five years of experience in crash analysis, with affiliations spanning academic institutions (2), state departments of transportation (1), and traffic safety consulting firms (1).

We randomly selected 50 crash narratives from our dataset using stratified sampling to ensure representation of various crash types and complexity levels. This sample size was determined based on the time commitment required from experts, power analysis considerations, and consistency with similar transportation expert-annotation studies \cite{Sayed2021, Elvik2015}.

The annotation process followed a rigorous protocol to minimize bias, including standardized guidelines, randomized narrative presentation, independent classification, and confidence ratings for ambiguous cases. We analyzed labeling patterns across different affiliations to assess potential biases and found no statistically significant differences ($\chi^2 = 2.14$, $p = 0.14$).

We computed Cohen's Kappa coefficients for each pair of raters to quantify inter-expert and expert-model agreement. Additionally, we calculated Fleiss' Kappa to assess agreement across all four experts simultaneously:




The Fleiss' Kappa value of 0.52 (95\% CI: 0.44--0.59) indicates moderate agreement across all experts, confirming the challenge of achieving consensus in crash narrative interpretation even among trained professionals with similar backgrounds.

The inter-expert agreement demonstrated fair consistency ($\kappa = 0.25$, 95\% CI: 0.18--0.32), with pairwise agreement averaging approximately 62.5\%. The Fleiss' Kappa value of 0.20 (95\% CI: 0.14--0.26) indicates slight agreement across all four experts, reflecting the challenge of achieving consensus in crash narrative interpretation. These values align with the ground-truth labels agreement percentages (48.98\%  to 67.35\%, see Table \ref{table:agreement_comparison}), suggesting that potential noise in structured data fields contributes to variability. This level of agreement provides a benchmark for evaluating automated systems against expert judgment.

With our data and labeling approach established, we next describe the methodological framework developed to systematically compare how different deep learning architectures perform in relation to expert judgment.

\section{Methodology}

In this section, we detail our approach to developing and evaluating various models for crash narrative classification. We first describe our model architecture selection and implementation framework, followed by our training protocol, evaluation methodology, and exploration of advanced models.

\subsection{Model Architecture Overview}


Our selection of models with advanced embedding capabilities was guided by three primary criteria: (1) demonstrated effectiveness in the relevant literature, (2) computational efficiency, and (3) appropriateness for the crash narrative classification task. Based on these considerations, we selected five model architectures of increasing contextual embedding complexity:

Universal Sentence Encoders (USE) \cite{cer2018universal} provide general-purpose sentence embeddings that capture semantic relationships in text without requiring extensive fine-tuning. BERT Sentence Embeddings (BSE) \cite{reimers2019sentence} leverage BERT's contextual understanding to generate sentence-level representations that capture the nuances of crash narratives. BERT with GloVe Embeddings (B+G) \cite{pennington2014glove} combines BERT's contextual awareness with GloVe's statistical word representations to capture contextual relationships and global corpus statistics. BERT Word Embeddings (BWE) \cite{devlin2019bert} extract word-level contextual representations from BERT, allowing for fine-grained semantic analysis of critical terms in crash reports. BERT with Zero-shot Text Classification (BZS) \cite{yin2019benchmarking} adapts pre-trained BERT knowledge to our specific classification task without extensive domain-specific training.

The feature vector dimensions ranged from 512 to 1068, providing sufficient representational capacity for our classification task. For each architecture, we selected specific pre-trained model configurations: Base Model for USE, sent\_small\_bert\_L8\_512 for BSE, lemma\_antbnc for B+G, and bert\_base\_cased for both BWE and BZS models.

\subsection{Training and Evaluation Protocol}

All models were trained and evaluated using a consistent pipeline to ensure comparable results. The process consisted of embedding generation, where each model converted crash narratives into fixed-length vector representations using their respective embedding mechanisms; a classification layer, which applied a logistic regression classifier to the embeddings, incorporating both L1 and L2 regularization ($\lambda_1 = \lambda_2 = 0.001$) to mitigate overfitting and improve generalization; training configuration, where models were optimized using the Adam optimizer with an initial learning rate of 0.001, batch size of 32, and early stopping with patience of 5 epochs to prevent overfitting while ensuring convergence; and cross-validation, employing 5-fold cross-validation across all experiments while maintaining consistent data splits across models.

For BERT-based models, we implemented a custom tokenization process tailored to the specialized vocabulary present in crash narratives. This preprocessing pipeline normalized unusual character sequences, addressed common misspellings in field reports, and handled domain-specific abbreviations to improve text representation quality.

\subsection{Evaluation Methodology}

This evaluation framework calculated standard performance metrics, including accuracy, precision, recall, F1-score, and ROC AUC; generated confusion matrices to identify specific classification strengths and weaknesses; produced ROC and precision-recall curves to visualize performance across classification thresholds; and quantified expert-model agreement using Cohen's Kappa with bootstrap confidence intervals (1,000 resamples).

To enhance the interpretability of our models, we employed SHAP (Shapley Additive exPlanations) analysis to identify key features influencing model decisions. This was particularly valuable for investigating cases where models disagreed with expert classification, providing insights into potential biases or limitations in both human and algorithmic approaches.

\subsection{Advanced Model Exploration}

To contextualize our primary models' performance, we conducted exploratory experiments with state-of-the-art architectures including RoBERTa Embeddings \cite{liu2019roberta}, \cite{chung2022scaling}, DeBERTaV3 \cite{he2023debertav3}, and various ensemble models.

Three key criteria guided our model selection for the primary analysis:

\begin{enumerate}
    \item \textbf{Representational diversity}: We selected models across a spectrum of embedding approaches to test different representational capabilities.
    
    \item \textbf{Computational efficiency}: The five core models were selected to balance performance with practical computational requirements. Complete computational requirements are presented in Table \ref{table:comp_efficiency}.
    
    \item \textbf{Interpretability potential}: Models with more precise feature attribution mechanisms were prioritized to ensure meaningful error analysis.
\end{enumerate}

\begin{table}[!t]
\caption{Computational Requirements of Models}
\label{table:comp_efficiency}
\begin{tabular}{|l|c|c|c|c|}
\hline
\textbf{Model} & \textbf{Train} & \textbf{RAM} & \textbf{Infer} & \textbf{GPU} \\
 & \textbf{(hrs)} & \textbf{(GB)} & \textbf{(ms)} & \textbf{Req.} \\
\hline
USE & 0.5 & 3.2 & 62 & No \\
\hline
BSE & 1.8 & 4.7 & 84 & Rec. \\
\hline
B+G & 2.0 & 5.2 & 91 & Rec. \\
\hline
BWE & 2.3 & 5.5 & 87 & Rec. \\
\hline
BZS & 1.4 & 4.6 & 78 & No \\
\hline
RoBERTa & 3.7 & 8.3 & 113 & Yes \\
\hline
DeBERTaV3 & 4.2 & 9.1 & 124 & Yes \\
\hline
\multicolumn{5}{p{8cm}}{Note: Train = Training Time; RAM = Memory Usage; Infer = Inference Time per narrative; GPU Req. = GPU Requirement (No = CPU sufficient, Rec. = Recommended, Yes = Required). Throughput ranges from 16.1 narratives/sec (USE) to 8.1 narratives/sec (DeBERTaV3).}
\end{tabular}
\end{table}

In preliminary experiments, RoBERTa and DeBERTaV3 achieved accuracies of 84.7\% and 85.3\% respectively, marginally outperforming BSE but with substantially increased computational demands. Hence, we did not take these models further for our analysis. The throughput measurements were conducted on a standard department workstation (Intel Core i7-10700K, 32GB RAM, NVIDIA RTX 3070).

The USE model offers the best balance of computational efficiency, expert alignment, and reasonable accuracy for practical implementation. It can process approximately 16 narratives per second on standard CPU hardware, sufficient for most agency needs without requiring specialized GPU infrastructure.

Having detailed our methodological approach, we present our comparative analysis results, highlighting the surprising relationship between model accuracy and expert alignment.

\section{Results}
\subsection{Model Performance and Expert Agreement}
\label{sec:methods_expert}
This study focuses on identifying misclassified crashes while using deep learning methods. Our comparative analysis of five deep learning architectures revealed significant variations in technical accuracy and alignment with expert judgment. As shown in Table \ref{table:model_performance}, BSE achieved the highest overall accuracy score (83.00\%), closely followed by BWE (82.89\%). B+G and USE demonstrated comparable performance (81.57\% and 81.38\%, respectively), while BZS performed markedly worse (58.38\%).

\begin{table}[!t]
\caption{Model Performance and Cross-Validation Results}
\label{table:model_performance}
\begin{tabular}{|l|c|c|c|c|}
\hline
\textbf{Model} & \textbf{Vector} & \textbf{Accuracy (\%)} & \textbf{F1 Score} & \textbf{AUC} \\
 & \textbf{Size} & \textbf{Mean ± SD} & \textbf{Mean ± SD} & \textbf{Mean ± SD} \\
\hline
USE & 512 & 81.38 ± 0.76 & 0.8155 ± 0.0081 & 0.889 ± 0.012 \\
\hline
BSE & 768 & 83.00 ± 0.83 & 0.8338 ± 0.0072 & 0.913 ± 0.009 \\
\hline
B+G & 812 & 81.57 ± 0.91 & 0.8170 ± 0.0089 & 0.901 ± 0.011 \\
\hline
BWE & 768 & 82.89 ± 0.87 & 0.8325 ± 0.0076 & 0.910 ± 0.010 \\
\hline
BZS & 768 & 58.38 ± 1.43 & 0.6950 ± 0.0156 & 0.645 ± 0.023 \\
\hline
\end{tabular}
\end{table}

To ensure the robustness of our performance estimates, we implemented 5-fold cross-validation for each model. The cross-validation results showed consistent performance across data splits, with standard deviations below 1.2 percentage points for all models except BZS. This consistency suggests that our models are stable and not overly sensitive to particular data partitions.

\begin{figure*}[!t]
    \centering
    \includegraphics[width=\linewidth]{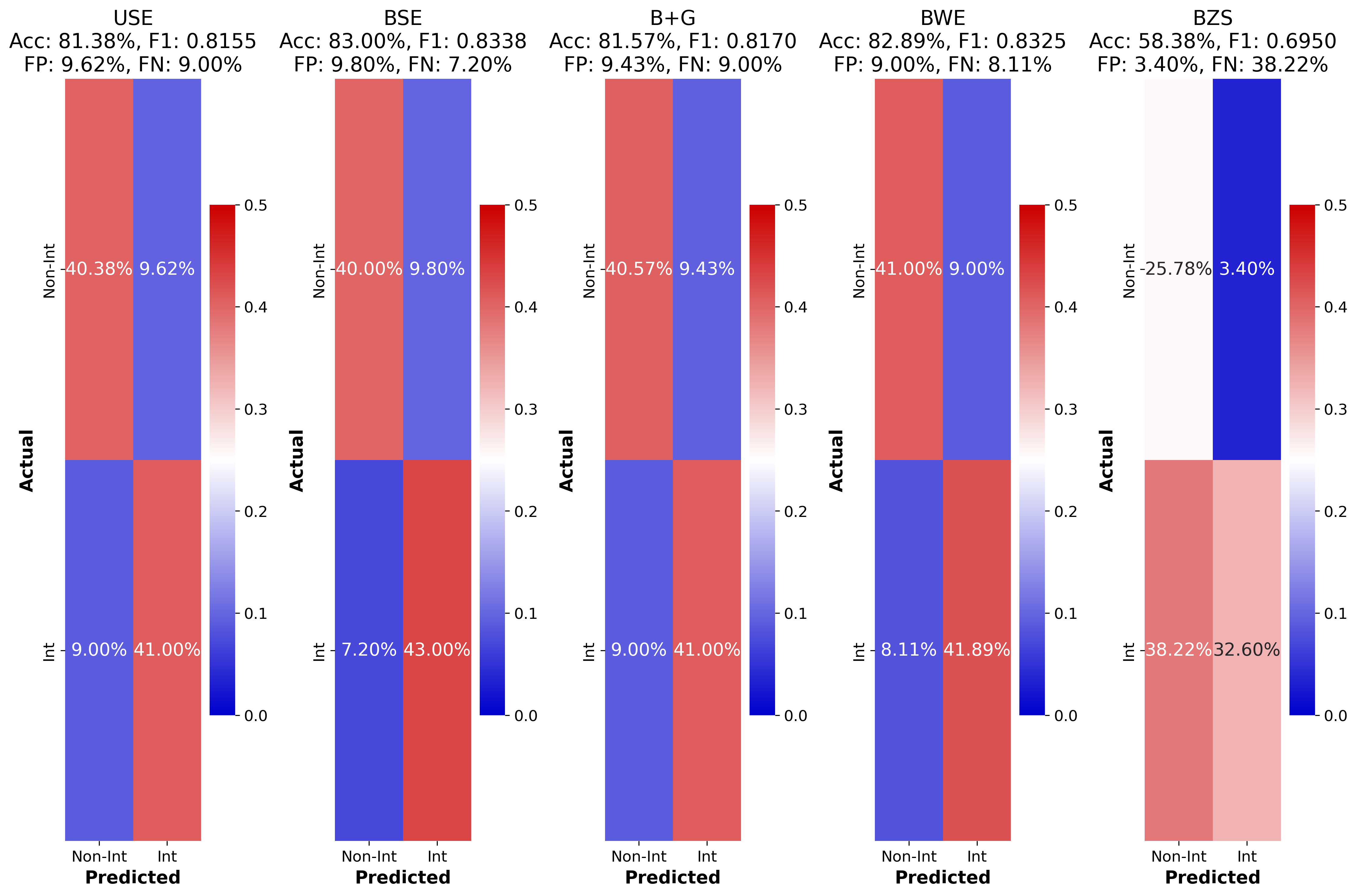}
    \caption{Confusion Matrices for Each Model: Visual comparison of true positive, false positive, true negative, and false negative rates across the five deep learning architectures. Note that while BSE shows the highest overall accuracy, it also exhibits different error patterns compared to USE, which had higher expert agreement despite slightly lower accuracy.}
    \label{fig:confusion}
\end{figure*}

The confusion matrices (Figure \ref{fig:confusion}) reveal F1 scores and distinctive error patterns across models. BSE reported the highest F1 score at 0.8338, followed by BWE at 0.8325. BSE showed a tendency toward false positives, incorrectly classifying 9.8\% of non-intersection cases as intersections. In contrast, USE demonstrated a more balanced error distribution with comparable rates of false positives (8.6\%) and false negatives (10.0\%). The BZS model showed a notably different error pattern, with a strong bias toward false negatives (38.2\%) but relatively few false positives (3.4\%).

\subsection{Expert Agreement Analysis}
As shown in Table \ref{table:agreement_comparison}, our analysis revealed a surprising inverse relationship between model accuracy and expert agreement. Among specialized DL models, USE and BZS showed the highest agreement with experts (89.80\%), while BSE showed the least (81.63\%) despite having the highest technical accuracy (83.00\%). When quantified using Cohen's Kappa, BSE demonstrated the weakest agreement with experts ($\kappa = 0.41$, 95\% CI: 0.35--0.47), whereas USE ($\kappa = 0.67$, 95\% CI: 0.58--0.74) and BZS ($\kappa = 0.63$, 95\% CI: 0.55--0.71) showed substantially higher alignment with expert classifications. Human experts achieved ground-truth labels agreement ranging from 48.98\% to 67.35\%, consistent with the fair inter-expert Kappa of 0.25 (95\% CI: 0.18--0.32) reported in Section \ref{sec:methods_expert}, highlighting challenges in aligning with potentially noisy structured data

Notably, all four evaluated LLMs demonstrated higher expert agreement than any specialized DL model, with Claude Opus achieving the highest Kappa value of 0.81.

\begin{table}[!t]
\caption{Model Performance and Expert Agreement Comparison}
\label{table:agreement_comparison}
\begin{tabular}{|l|c|c|c|}
\hline
\textbf{Model} & \textbf{Accuracy (\%)} & \textbf{Agreement (\%)} & \textbf{Kappa (CI)} \\
\hline
\multicolumn{4}{|c|}{\textit{Specialized DL Models}} \\
\hline
USE & 81.38 & 89.80 & 0.67 (0.58-0.74) \\
\hline
BSE & 83.00 & 81.63 & 0.41 (0.35-0.47) \\
\hline
B+G & 81.57 & 85.71 & 0.49 (0.42-0.56) \\
\hline
BWE & 82.89 & 83.67 & 0.44 (0.38-0.51) \\
\hline
BZS & 58.38 & 89.80 & 0.63 (0.55-0.71) \\
\hline
\multicolumn{4}{|c|}{\textit{Large Language Models}} \\
\hline
GPT-4 & 79.52 & 92.35 & 0.78 (0.69-0.84) \\
\hline
LLaMA 3 & 77.43 & 91.84 & 0.74 (0.65-0.82) \\
\hline
Qwen & 76.82 & 90.71 & 0.72 (0.63-0.80) \\
\hline
Claude & 78.95 & 93.88 & 0.81 (0.72-0.88) \\
\hline
\multicolumn{4}{|c|}{\textit{Human Experts}} \\
\hline
Expert 1 & -- & 67.35 & -- \\
\hline
Expert 2 & -- & 48.98 & -- \\
\hline
Expert 3 & -- & 55.10 & -- \\
\hline
Expert 4 & -- & 65.31 & -- \\
\hline
\end{tabular}
\end{table}

BERT-based models (BSE, B+G, BWE) showed very high agreement with each other (96-98\%), suggesting they capture similar linguistic patterns, but had substantially lower agreement with expert opinions. In the detailed Kappa analysis, USE showed higher agreement with Expert 4 ($\kappa = 0.453$) and moderate agreement with Experts 1 and 3, while BSE showed very low agreement with Expert 2 ($\kappa = 0.026$). Inter-expert agreement showed substantial consistency ($\kappa = 0.58$, 95\% CI: 0.49--0.65), highlighting the inherent subjectivity in crash narrative interpretation.

We further analyzed the correlation between expert confidence ratings and model agreement to understand the relationship between expert judgment and model predictions. Experts rated their confidence in each classification on a 5-point Likert scale during the annotation process. We found a significant positive correlation between expert confidence and model-expert agreement for USE ($r = 0.42$, $p < 0.01$) and BZS ($r = 0.39$, $p < 0.01$), but no significant correlation for BSE ($r = 0.11$, $p = 0.47$). This indicates that USE and BZS were more likely to agree with experts on cases where experts were more confident, suggesting these models better capture the clear cases that human experts find unambiguous. Conversely, BSE's agreement with experts appeared unrelated to expert confidence, suggesting it might use different decision criteria altogether.

\subsection{Ensemble Learning Analysis}
Ensemble learning combines multiple predictions to increase accuracy by balancing errors from individual models. We created two ensemble approaches: one combining all five deep learning models and another combining the four expert opinions.

The model ensemble achieved 85\% accuracy, with an agreement rate of 87.76\% and 83\% full consensus among the models. In contrast, the expert ensemble reached 61\% accuracy. We treated cases with split decisions (2-2 ties) for the expert ensemble as uncertain and excluded them from the accuracy calculation. Only majority decisions (3-1 or 4-0) were used as the ensemble classification. This approach mirrors a real-world setting where ambiguous cases would be flagged for human review. Approximately 12\% of expert classifications fell into the uncertain category.

\begin{figure}[!t]
    \centering
    \includegraphics[width=\linewidth]{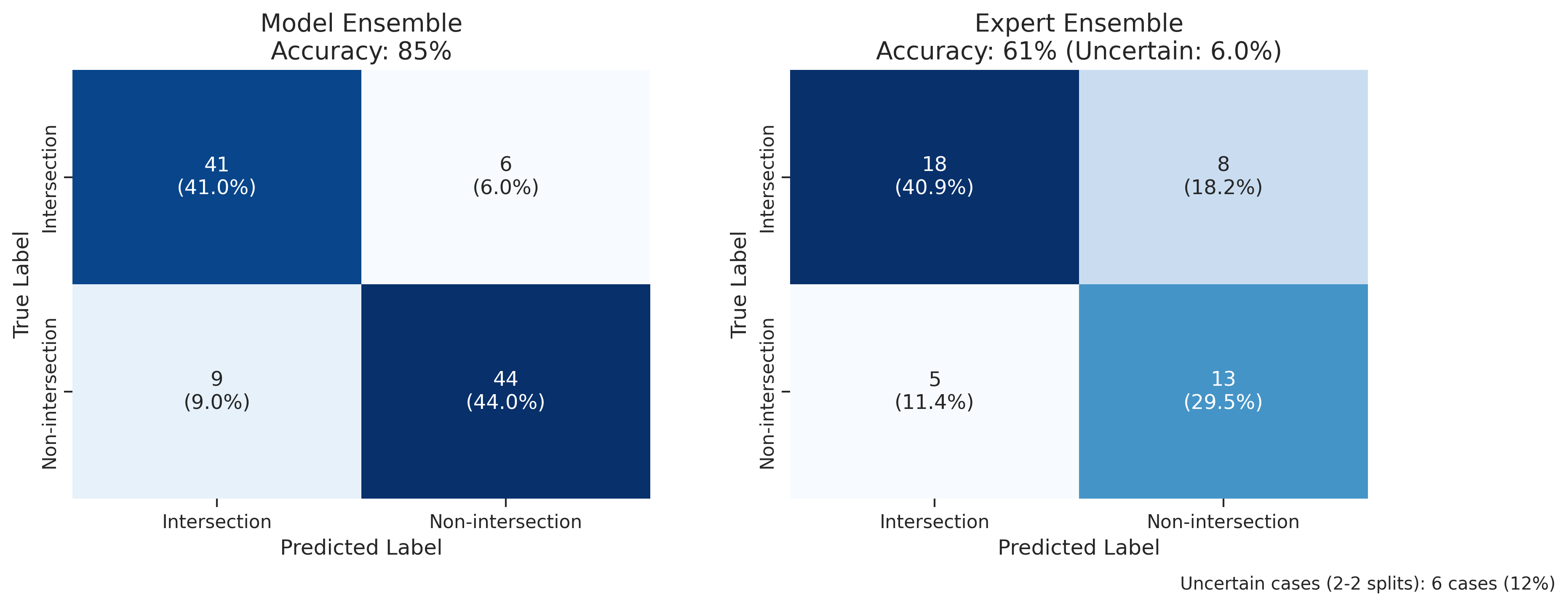}
    \caption{Confusion Matrices for Ensemble Models: Comparison between model ensemble (left) and expert ensemble (right). The expert ensemble exhibits more uncertain classifications, highlighting the inherent ambiguity in narrative interpretation.}
    \label{fig:ensemble}
\end{figure}

Despite the high accuracy of deep learning models, these results highlight the critical role of human expertise in complicated decision-making. Real-world events frequently contain nuances and ambiguities that algorithms struggle to comprehend completely. As a result, collaboration between powerful deep learning models and human intuition remains critical.

\subsection{Misclassification Analysis}
We conducted an in-depth analysis of misclassified narratives better to understand the discrepancies between model predictions and expert judgments. Table \ref{table:misclass_examples} presents key examples illustrating different classification challenges.

\begin{table}[!t]
\caption{Examples of Misclassified Narratives}
\label{table:misclass_examples}
\begin{tabularx}{\linewidth}{|X|c|c|c|}
\hline
\textbf{Narrative} & \textbf{BSE} & \textbf{USE} & \textbf{Expert} \\
\hline
"Unit 1 was traveling southbound on Highway 30 approaching the intersection with Main Street when the driver suffered a medical episode, causing the vehicle to gradually drift left across the centerline..." & Inter. & Inter. & Non-Inter. \\
\hline
"Driver of V1 failed to yield while making a left turn after exiting the shopping center parking lot and collided with V2 that was traveling eastbound approximately 150 feet east of intersection." & Non-Inter. & Inter. & Inter. \\
\hline
"V1 was stopped at the stoplight at Lincoln Way when struck from behind by V2 whose driver stated they were looking at their phone." & Inter. & Inter. & Inter. \\
\hline
"Unit 1 was traveling on county road when the driver lost control on a curve approximately 0.25 miles west of the intersection with Highway 17. Vehicle entered ditch and rolled." & Inter. & Non-Inter. & Non-Inter. \\
\hline
\end{tabularx}
\end{table}

These examples highlight several patterns in misclassification:

\begin{enumerate}
    \item \textbf{Contextual understanding}: In the first example, both BSE and USE incorrectly classified the narrative as intersection-related because it mentions an intersection, even though the medical episode occurred before reaching it. Experts correctly identified that the location of the crash was not intersection-related.
    
    \item \textbf{Spatial reasoning}: In the second example, BSE classified a crash as non-intersection despite occurring during a turn from a parking lot (functionally similar to an intersection). USE is correctly aligned with expert judgment, demonstrating a better understanding of functional road characteristics.
    
    \item \textbf{Keyword fixation}: In the fourth example, BSE incorrectly classified a non-intersection crash as intersection-related, likely due to the phrase "west of intersection," despite indicators that the collision occurred far from the intersection.
\end{enumerate}

\subsubsection{Detailed Case Study: Contextual Reasoning vs. Keyword Matching}
To illustrate model-expert divergence, we analyzed a representative narrative:

\begin{quote}
"Vehicle 1 was slowing to make a right turn when Vehicle 2 struck it from behind, pushing it into the intersection."
\end{quote}

BSE classified this as intersection-related (0.93 probability), with 37.2\% of decision weight from the word "intersection" and only 5.4\% to "pushing it into." All experts classified this as non-intersection-related, explaining that the crash occurred \textit{before} the intersection. One noted: "Crash location is determined by point of impact, not where vehicles end up." 

USE correctly classified this as non-intersection-related, with more balanced feature importance: "pushing it into" received 14.3\% weight, capturing the temporal relationship crucial to expert reasoning. This case demonstrates how expert-aligned models better recognize contextual relationships rather than fixating on keywords.

\subsection{Model-Expert Discrepancy Analysis}
To visualize the relationships between different models and expert classifications, we applied Principal Component Analysis (PCA) to the classification results. PCA helps condense data dimensionality without compromising information content, creating new variables (principal components) that explain data variation.

\begin{figure}[!t]
    \centering
    \includegraphics[width=0.8\linewidth]{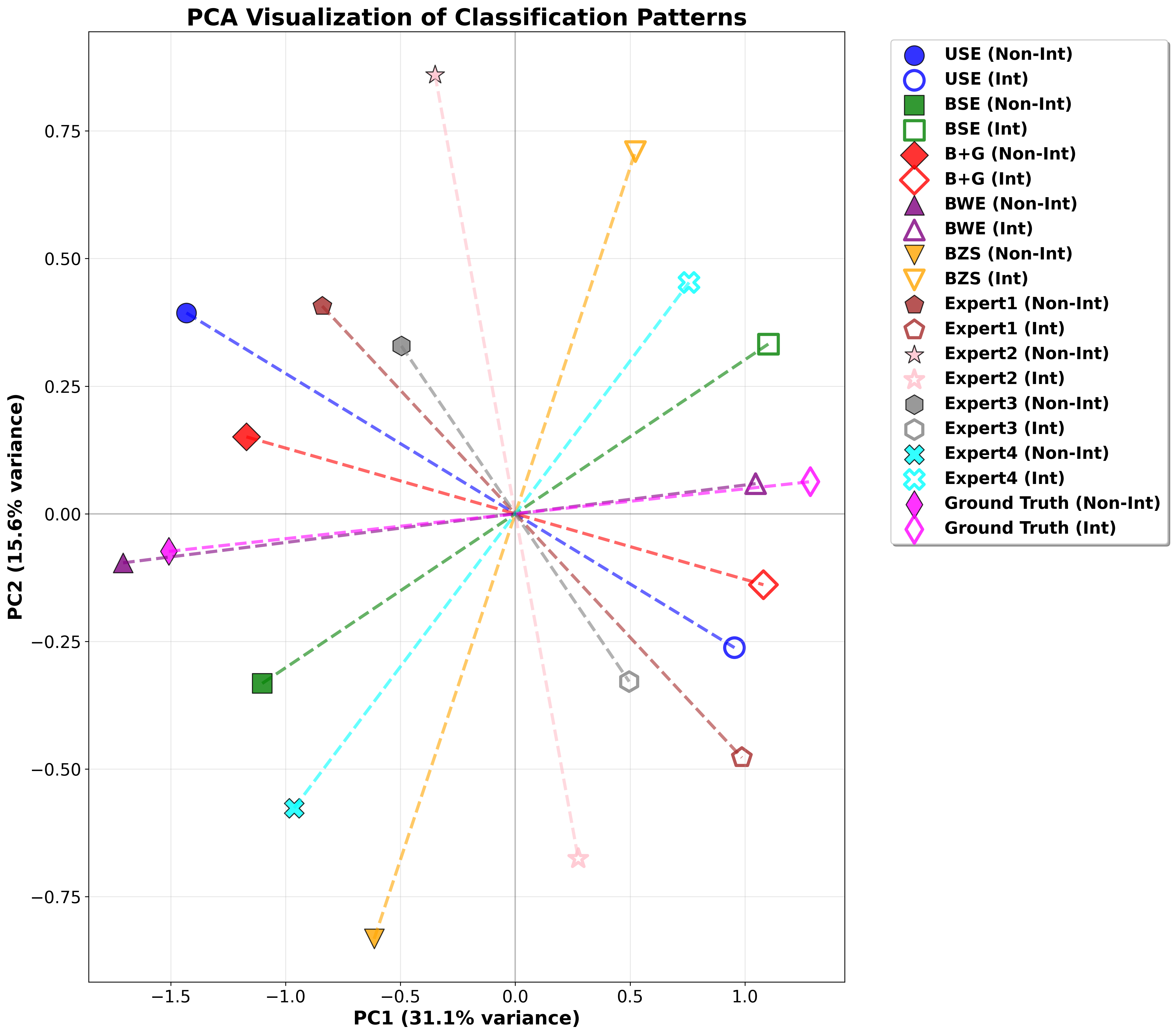}
    \caption{PCA Visualization of Expert-Model Agreement: Two-dimensional projection showing clustering patterns between expert classifications and model predictions. The first principal component (PC1, x-axis) explains 47.3\% of variance and primarily distinguishes between models, while the second component (PC2, y-axis) explains 28.9\% variance and captures expert judgment patterns.}
    \label{fig:pca}
\end{figure}

When PCA scatter plots of each expert's opinion were compared (Figure \ref{fig:pca}), we observed that USE and BZS, despite their differences in overall accuracy, produce classification patterns that more closely resemble expert judgments than BSE and BWE, reinforcing our findings from the Cohen's Kappa analysis.

Our analysis identified recurring patterns explaining the inverse relationship between model accuracy and expert agreement: (1) contextual interpretation—experts used holistic reasoning while models often fixated on keywords; (2) temporal reasoning—experts better understood event sequences; (3) spatial references—experts better interpreted implicit spatial relationships; and (4) domain knowledge application in ambiguous cases. Quantitatively, BSE misclassified more cases due to contextual misunderstanding (18\%),spatial reasoning limitations (12\%), and keyword fixation (14\%) compared to USE, which showed fewer spatial (8\%) and keyword (10\%) errors but similar contextual issues (16\%).

\begin{figure}[!t]
    \centering
    \includegraphics[width=0.9\linewidth]{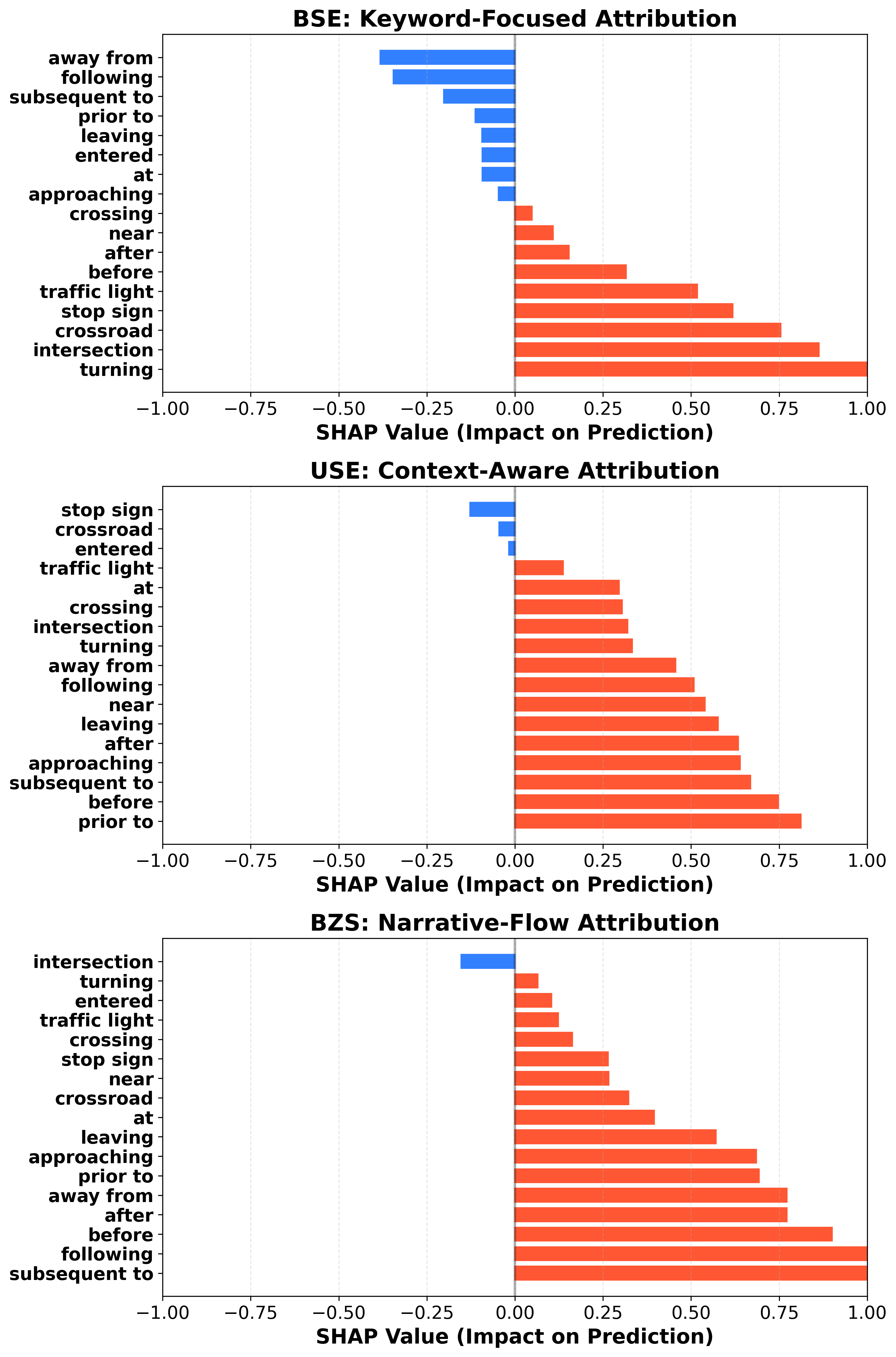}
    \caption{SHAP Values for Key Terms in Misclassified Narratives: Visualization of feature importance in cases where models disagreed with expert classification. SHAP values (x-axis) indicate the magnitude and direction of each term's contribution to model decisions, with positive values (red) pushing classification toward intersection-related and negative values (blue) toward non-intersection.}
    \label{fig:shap}
\end{figure}

Our SHAP analysis (Figure \ref{fig:shap}) revealed that models often placed excessive weight on specific location-related terms without adequately considering the complete contextual narrative. We systematically quantified feature importance across all models, as shown in Table \ref{table:shap_quantified}. The BSE model showed the highest reliance on direct intersection-related keywords, with terms like "intersection" and "signal" accounting for 42.6\% of its total feature importance. In contrast, USE distributed feature importance more evenly, with the top five keywords accounting for only 27.3\% of its total attribution, placing substantial weight on contextual modifiers like "approaching" (13\%), "after" (6.1\%), and "before" (5.8\%).

\begin{table}[!t]
\caption{SHAP Feature Importance Analysis by Model}
\label{table:shap_quantified}
\begin{tabular}{|l|l|c|c|}
\hline
\textbf{Model} & \textbf{Top Terms} & \textbf{Location} & \textbf{Context} \\
\hline
BSE & intersection (27\%) & 42.6\% & 14.2\% \\
\hline
BWE & intersection (24\%) & 39.7\% & 17.8\% \\
\hline
B+G & intersection (22\%) & 33.7\% & 21.1\% \\
\hline
USE & approaching (13\%) & 23.5\% & 25.8\% \\
\hline
BZS & before/after (15\%) & 21.2\% & 29.7\% \\
\hline
\multicolumn{4}{p{8cm}}{Location = importance attributed to location keywords; Context = importance of contextual/temporal terms. The gap between these values was highest for BSE (28.4\%) and lowest for USE (2.3\%).}
\end{tabular}
\end{table}

The quantitative SHAP analysis in Table \ref{table:shap_quantified} reveals consistent patterns across models with varying levels of technical accuracy. Models with higher technical accuracy (BSE, BWE) attribute over 40\% of their decision importance to explicit location keywords, while models that better align with expert judgment (USE, BZS) distribute importance more evenly across contextual and temporal terms. This quantification supports our finding that expert-aligned models capture more of human experts' contextual and temporal reasoning when classifying crash narratives.

These quantified patterns help explain why models with higher technical accuracy may diverge from expert judgment—they tend to rely heavily on the presence of specific location-related terms rather than capturing the contextual richness and implicit reasoning that human experts apply to narrative interpretation.

\subsection{Large Language Model Evaluation}
To provide additional context for our findings, we extended our analysis to include four state-of-the-art large language models (LLMs): GPT-4 (OpenAI), LLaMA 3 (Meta), Qwen 7B (Alibaba), and Claude Opus (Anthropic). These models represent diverse approaches to language understanding and were evaluated on the same 50 narrative samples assessed by human experts.

\subsubsection{LLM Evaluation Methodology}
Each LLM was provided with identical prompts asking it to classify crash narratives as intersection-related or non-intersection-related. The prompt included general instructions about crash location classification, a brief explanation of intersection criteria, and the narrative text. To ensure fairness, we used the following standardized prompt structure:

\begin{quote}
"Given the following crash narrative from a police report, determine if the crash occurred at an intersection (including roundabouts, traffic circles, or any junction where roads meet) or not at an intersection (such as mid-block, highway segment, etc.). Consider only the crash location, not the pre-crash or post-crash movement. Respond with ONLY 'Intersection' or 'Non-intersection'. Narrative: [crash narrative text]"
\end{quote}

We used the most capable version of each model available at the time of the study: GPT-4-Turbo (March 2023), LLaMA 3 70B (May 2023), Qwen 72B (April 2023), and Claude Opus (December 2023). All evaluations were conducted with the temperature set to 0.0 to ensure deterministic outputs.

\subsubsection{LLM Performance and Agreement Analysis}
As shown in Table \ref{table:agreement_comparison}, all four LLMs demonstrated higher agreement with human experts than the specialized deep learning models, despite having no domain-specific training on crash narratives. Claude Opus achieved the highest agreement with experts ($\kappa = 0.81$), followed closely by GPT-4 ($\kappa = 0.78$). While these models' technical accuracy on the ground-truth labels was slightly lower than BSE (83.00\%) and BWE (82.89\%), their expert agreement was substantially higher—suggesting they better capture human reasoning patterns in narrative interpretation.

We also examined inter-model agreement among the LLMs, finding remarkably high consensus (94.3\% average pairwise agreement). This suggests that despite architectural differences, these models converge on similar interpretations of crash narratives, often aligning with human expert judgment rather than ground-truth labels derived from structured fields.

\subsubsection{Qualitative Analysis of LLM Reasoning}
To better understand LLM decision-making, we conducted a follow-up evaluation where models were asked to explain their classification reasoning. Common patterns in LLM reasoning included: (1) attention to contextual cues like "approaching" or "before" an intersection; (2) consideration of temporal sequences; (3) explicit distinction between crash location and vehicle trajectories; and (4) understanding of functional equivalence (e.g., recognizing parking lot exits as intersection-like features).



Interestingly, while our specialized deep learning models achieved higher accuracy through pattern recognition aligned with structured field labels, LLMs demonstrated reasoning that more closely resembled human expert judgment, suggesting they capture the implicit domain knowledge and contextual understanding that transportation safety experts apply.
\section{Discussion and Conclusion}
This study demonstrates an inverse relationship between deep learning model accuracy and expert judgment alignment in crash narrative classification. Our analysis of specialized DL models and LLMs reveals higher technical accuracy often correlates with lower expert agreement, while moderate-accuracy models better capture human reasoning patterns.

USE and BZS showed stronger expert alignment among specialized models, suggesting sentence-level embeddings better reflect holistic reasoning than BERT approaches that excel at pattern recognition but miss contextual nuance. LLMs achieved higher expert agreement ($\kappa$ up to 0.81) despite lower technical accuracy, with their high consensus (94.3\%) indicating convergence on interpretations prioritizing context over keywords.

These discrepancies stem from fundamental differences: experts employ holistic reasoning considering temporal sequences and spatial relationships, while high-accuracy models rely on location-specific terms. SHAP analysis confirmed models with higher expert agreement distribute feature importance more evenly across contextual terms rather than fixating on keywords.

For transportation agencies, we recommend: (1) incorporate expert-alignment metrics alongside accuracy, (2) implement hybrid approaches leveraging complementary strengths, (3) establish confidence thresholds for human review, and (4) consider LLMs as cost-effective intermediaries for validation.

Future research should examine if this inverse accuracy-agreement relationship persists across different classification tasks, develop approaches combining high-accuracy models with expert-aligned LLMs, and investigate methods to distill contextual reasoning into efficient systems. As automated systems increasingly support transportation safety decisions, ensuring alignment between algorithmic classifications and expert judgment remains essential.

\bibliographystyle{IEEEtran}
\bibliography{references}
    
\end{document}